\documentclass[10pt,letterpaper]{article}

\usepackage{cogsci}
\cogscifinalcopy

\usepackage{pslatex}
\usepackage{float}
\usepackage{qtree}
\usepackage{graphicx}
\usepackage{adjustbox}

\usepackage{natbib}

\usepackage{amsmath}
\newcommand{\varA}[1]{{\operatorname{\mathit{#1}}}}

\usepackage[hidelinks]{hyperref}

\title{Anaphoric Structure Emerges Between Neural Networks}

\author{
\large{ \bf Nicholas Edwards, Hannah Rohde$^{\circ}$, Henry Conklin$^{\circ,\bullet}$}\\
\emph{ne283@cantab.ac.uk, \{hannah.rohde, henry.conklin\}@ed.ac.uk} \\
$^{\circ}$Department of Linguistics \& English Language\\
$^{\bullet}$School of Informatics\\
The University of Edinburgh}

\begin{document}

\maketitle

\begin{abstract}
Pragmatics is core to natural language, enabling speakers to communicate efficiently with structures like ellipsis and anaphora that can shorten utterances without loss of meaning. These structures require a listener to interpret an ambiguous form—like a pronoun—and infer the speaker's intended meaning—who that pronoun refers to. Despite potential to introduce ambiguity, anaphora is ubiquitous across human language. In an effort to better understand the origins of anaphoric structure in natural language, we look to see if analogous structures can emerge between artificial neural networks trained to solve a communicative task. We show that: first, despite the potential for increased ambiguity, languages with anaphoric structures are learnable by neural models. Second, anaphoric structures emerge between models `naturally' without need for additional constraints. Finally, introducing an explicit efficiency pressure on the speaker increases the prevalence of these structures. We conclude that certain pragmatic structures straightforwardly emerge between neural networks, without explicit efficiency pressures, but that the competing needs of speakers and listeners conditions the degree and nature of their emergence.

\end{abstract}

\section{Introduction}
When we communicate, we often leave out material that is recoverable from the context. Linguistically, such omission or shortening presents a challenge for the listener—both in identifying that something has been omitted and recovering the intended meaning in context. Certain linguistic structures signal that part of what's being said is redundant and can be recovered: ellipsis enables speakers to signal repeated meaning by omitting words \citep{rooth1992ellipsis} and pronominal anaphora signals the re-mention of a discourse referent (see Figure~\ref{egfig} for examples). Both structures employ \emph{anaphors} that refer back to meaning mentioned elsewhere in the context—\emph{antecedents}. While this is particularly clear in the case of pronouns, work on ellipsis has suggested that there is similar anaphoric behaviour at ellipsis sites \citep{hankamer_deep_1976,sag_deletion_1976,hardt1993verb,fiengo_indices_1994}, even when the missing material is not replaced with an overt marker, like in cases of null anaphora, or \emph{pro-drop} \citep{chomsky_lectures_1981}.  
We take both pronouns and ellipsis as examples of the broader class of pragmatic structures in natural language. We look at what conditions are needed for structures analogous to these to arise between neural networks in an effort to better understand why pragmatic structure may have emerged across human languages despite its potential for ambiguity.

\begin{figure}
\begin{center}
\begin{tabular}{ll}

$\updownarrow$ & \begin{tabular}{ll}
{\fontfamily{cmss}\selectfont \scriptsize meaning:} & $dances'(John) \land~dances'(Mary)$\\
{\fontfamily{cmss}\selectfont \scriptsize form:} & John dances and Mary does too. 
\end{tabular}\\[4ex]

\hline \\

$\updownarrow$ & \begin{tabular}{ll}
{\fontfamily{cmss}\selectfont \scriptsize meaning:} & $sits'(Mary) \land~reads'(Mary)$ \\
{\fontfamily{cmss}\selectfont \scriptsize form:} & Mary sits and (she) reads. 

\end{tabular}\\

\end{tabular}
\end{center}
\caption{Examples of sentences with anaphoric structure used in our experiments. The first example illustrates \emph{verb phrase ellipsis}, where the repetition of verb meaning is signalled by \emph{does too}. The second example illustrates \emph{pronominal anaphora}, where the pronoun \emph{she} can be used to signal the re-mention of a previous discourse referent, \emph{Mary}. In some languages, this pronoun can be omitted.}
\label{egfig}
\end{figure}

A growing body of work makes the case that natural language has evolved to enable efficient communication between humans, with the competing needs of speakers and listeners as major factors shaping the structure that emerges \citep*{hawkins2004efficiency,jaeger2011language,piantadosi_word_2011,kemp2018semantic}. This is perhaps most famously demonstrated by Zipf’s Law \citep{zipf1949human}, which observes that word frequency is inversely correlated with word length: more frequent words (e.g., \emph{a}) are shorter than less frequent words (\emph{e.g., electroencephalograph}), which helps minimise the production effort required by a speaker \citep{macdonald2013language}. Compressing semantically redundant information could be argued to achieve a similar goal. However, compression risks introducing greater ambiguity into communication, increasing a listener's uncertainty about the intended meaning \citep*{piantadosi_communicative_2012}. Despite such a risk of miscommunication, the existence of these structures across the world's languages \citep{huang2000anaphora} suggests that affordances to the speaker outweigh potential communicative failure. Taken together, these observations illustrate how language balances the needs of the speaker by minimising the costs of production while allowing the listener to recover the meaning behind what is said \citep{levinson2000presumptive,macdonald2013language,gibson2019efficiency}.

\begin{figure}
\begin{center}
\includegraphics[width=0.48\textwidth]{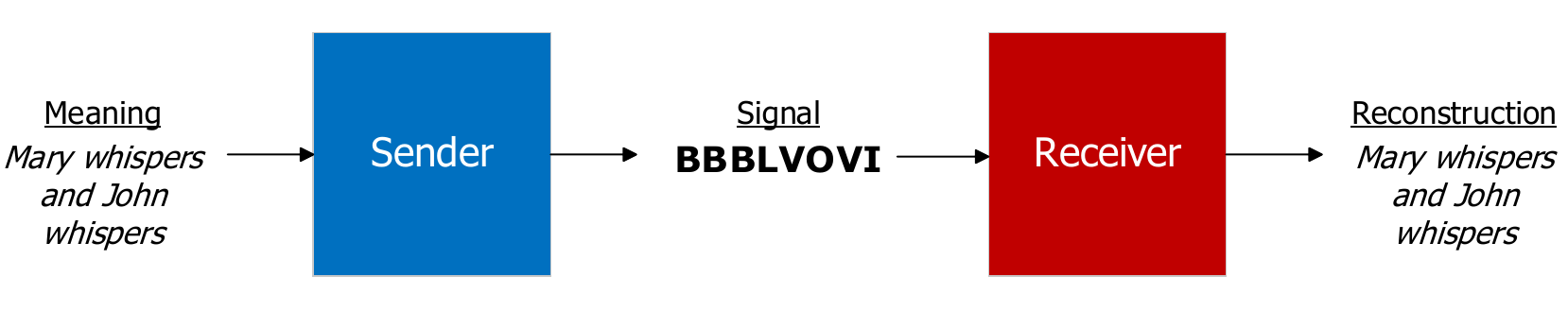}
\end{center}
\caption{Illustration of the Lewisian signalling game. The Sender generates a signal representing the meaning, and the Receiver guesses the original meaning by decoding the signal. Here, the Receiver correctly reconstructs the meaning, so the round is successful.} 
\label{setupfig}
\end{figure}

Recent work also looks at languages that emerge between neural agents trained to solve a communicative task \citep{kottur_natural_2017,lazaridou2018emergence}. The task is modelled after a Lewisian signalling game \citep{lewis2008convention} (Figure \ref{setupfig}) where agents need to communicate about a meaning space but are given no supervision about how to do so. During a run of the model, a language which maps meanings to signals emerges, enabling communication between agents. Consequently, the languages are shaped by the biases of the networks and the objective with which they're trained \citep{lazaridou_emergent_2020}.
Recent work in this area has largely been concerned with investigating and identifying the conditions required for the emergence of syntactic structure \citep{kottur_natural_2017,mordatch_emergence_2018,ren2020compositional,conklincompositionality}. Other work has examined contact linguistics, showing, for example, that creole-like languages can emerge in populations of agents tasked with playing a simple reference game (Graesser et al., 2019). In our work, we instead study the emergence of pragmatic structure through the lens of multi-agent communication, which has so far received little attention from the emergence and efficient communication literature. We consider whether anaphoric structure analogous to that found in human language can emerge and what conditions might be required for this to happen. Neural networks have been shown to lack human inductive biases on an array of linguistic tasks \citep*{chaabouni_anti-efficient_2019,mccoy2019right,conklin-etal-2021-meta}, which makes them an interesting testing ground for efficiency-driven accounts of anaphora. If anaphoric structure can emerge in a population of speakers that don't share our cognitive endowment, this could support the idea that these structures emerge predominantly as a result of pressures arising from the competing communicative needs of speaker and listener.

Prior work investigating signal compressibility at the language level \citep{chaabouni_anti-efficient_2019} found that networks by default preferred signals that were \emph{anti}-efficient, where encodings of more frequent meanings were longer rather than shorter. Only when an explicit cost encouraging brevity was introduced did signals conform to Zipf’s Law. This suggests that the agents opted for a strategy which maximised listener discrimination, with the setup skewing in favour of listener-oriented pressures by default. In both that work and ours, the objective used to train the model optimises for as little ambiguity as possible in the listener's reconstruction of a meaning. With that in mind, we similarly look to see if anaphoric structure can emerge in the default (anti-efficient) case, or if an explicit efficiency pressure is required.

We put forward a set of quantitative measurements designed to capture three high-level characteristics of anaphoric structure, so we can identify if and when it emerges, given that the emerging structures may not necessarily appear as straightforward `she' or `did too' anaphora. Our measures assess: the uniqueness of structures used to signal meaning redundancy (`signal uniqueness'), signal ambiguity, and signal length. Before looking at whether these structures emerge between neural agents, we first train a single `listener' agent on handcrafted languages designed to mimic three attested types of anaphoric structures in natural language. We show that all three types of structures are learnable, but differ in the speed at which they're learned and the degree of ambiguity they impose on the listener. Then, in a set of multi-agent experiments we show that structures akin to anaphora in natural language emerge between neural agents in every run of every condition of our model—even without an explicit efficiency pressure on the `speaker' agent. By introducing an efficiency pressure, we increase the degree of anaphoric structure that emerges. None of the languages that emerge show substantial evidence of elided structures like those seen in pro-drop languages, which reaffirms that models like ours are optimised for minimal ambiguity. Taken together, our results suggest that while efficiency pressures on a speaker condition the anaphoric structure that emerges, such structure can emerge wherever there's redundancy in what's communicated. Such a finding points to the importance of the semantics-pragmatics interface, in addition to communicative needs, in providing an account of the origins of anaphoric structure.

\section{Methodology}
\subsection{The Game}
We use a \emph{reconstruction} game in the style of \cite{lewis2008convention}, with a Sender (`speaker' agent) and a Receiver (`listener' agent)—both neural networks. In each round:
\begin{enumerate}
    \item The Sender receives a meaning $m_i \in M$, drawn from the meaning space $M$, as input.
    \item The Sender generates a signal $s_i$ of maximum length $n$, one character $c$ at a time from an alphabet $C$ of size $|C|$.
    \item The Receiver receives the Sender's signal $s_i$ as input and predicts the corresponding meaning $\hat{m_i}$.
    \item  The round is successful if  $m_i = \hat{m_i}$.
\end{enumerate}
    
\subsection{The Meaning Space}
To allow for meanings with repetition, each meaning $m_i$ is the concatenation of 5 $\langle role, word \rangle$ pairs representing a `sentence', where each role represents an element of the sentence, like a subject or verb, and can be realized as a particular word:  e.g., $\langle subj_1, John\rangle$, $\langle verb_1, walks\rangle$, $\langle conj, and\rangle$, etc. Each sentence is grammatically structured as two conjoined one-place predicates with roles $(subj_1, verb_1, conj, subj_2, verb_2)$ as in: $walks'(John)$ $\land$ $smiles'(Mary)$. This yields three different kinds of redundant meanings:

\begin{itemize}
    \item{\textbf{Non-redundant:} nothing is repeated—e.g., \emph{John walks and Mary smiles} \\ $(subj_1 \neq subj_2 \land verb_1 \neq verb_2)$}.
    
    \item \textbf{Partially Redundant:} either the subject or verb is repeated—e.g., \emph{Mary walks and Mary smiles} \\
    $(subj_1 = subj_2 \lor verb_1 = verb_2)$.
    
    \item{\textbf{Fully Redundant:} if both the subject and verb are repeated—e.g., \emph{Mary smiles and Mary smiles} \\
    $(subj_1 = subj_2 \land verb_1 = verb_2)$}.
\end{itemize}

\noindent Here each $role$ is realised as one of 15 $words$—apart from the conjunction which can only be `and'—resulting in a meaning space size of 20,000. We use the same meaning space for both the single-agent and multi-agent experiments.

\subsection{Implementation}
The game is implemented in PyTorch \citep{paszke2019pytorch}, using portions of code from the EGG (`Emergence of lanGuage in Games') toolkit \citep{kharitonov_egg_2019}. The Sender is comprised of a linear layer which maps the meaning to a hidden representation of size 250 and a single-layer Gated Recurrent Unit (GRU) \citep{cho2014properties} that produces the variable-length signal a character at a time. The Receiver architecture is the inverse with a GRU mapping the signal to a hidden representation, and a linear layer mapping that to a predicted meaning. The Sender is limited to a maximum length, but may produce a signal of any length up to that bound.

\subsection{Optimisation}
A hybrid approach is used to train the agents. Since the loss is differentiable, the Receiver can be trained using standard stochastic gradient descent. Due to the discrete signal, the Sender is trained using policy gradient method REINFORCE \citep{williams_simple_1992}. Both are optimised using the Adam optimiser \citep{kingma2014adam}, with learning rate 0.001.\footnote{Our code and data, along with a full set of hyperparameters, can be found at: \url{https://github.com/hcoxec/emerge}.}

\section{Neural agents can learn languages with anaphoric structure}
Before seeing if anaphoric structure emerges between agents, we start by showing that the agents used in the referential game are capable of learning these structures if presented with a pre-existing language containing them. We design three languages for comparison, and train only the Receiver via supervised learning to map signals to meanings. The languages are:

\begin{enumerate}
    \item \textbf{No Elision}: each meaning role is mapped to one or more characters in the signal, even when there is redundancy.
    \item \textbf{Pronoun}: redundant roles are mapped to a unique anaphoric token—repeated nouns to a `pronoun' character, and repeated verbs to a `did too' character.
    \item \textbf{Pro-drop}: whenever there is a redundant role, the corresponding signal is \emph{not} appended, shortening the overall signal. The language is named after the similar linguistic phenomenon of \emph{pro-drop} \citep{chomsky_lectures_1981}.
\end{enumerate}

\noindent We use a miniature Receiver—given the simplicity of the task—with a hidden state of size 64, trained on each language for 50 epochs with a learning rate of $5\mathrm{e}^{-4}$. Results are averaged over 10 runs and summarised in Table~\ref{epochacc}.

\begin{table}[H]
    \centering
    \begin{tabular}{c | c | c}
    No Elision & Pronoun & Pro-drop \\
    \hline
    11.3 & 15.3 & 24.3 \\
    \end{tabular}
    \caption{Average number of epochs to reach 100\% accuracy on test set.}
    \label{epochacc}
\end{table}

All three languages are consistently learned by the Receiver, with the No Elision language learned faster than Pronoun ($t(18)=-4.35$, $p<0.05$), and the Pronoun language learned faster than Pro-drop ($t(18)=-6.19$, $p<0.05$).
This finding means the Receiver can successfully learn that in the signals corresponding to \emph{Mary smiles and she dances} and \emph{Ada smiles and she dances} the same token `she' maps to different nouns: Mary and Ada. 
\begin{figure}
\begin{center}
\begin{tabular}{ll}
\begin{tabular}{ll}
\emph{Sentence:}
\end{tabular}
& \begin{tabular}{l} John smiles and John smiles.
\end{tabular}\\
\hline
\begin{tabular}{ll}
\emph{Logical Form:}
\end{tabular}
& \begin{tabular}{l} $smiles'(John) \land~smiles'(John)$
\end{tabular}\\
\hline
\begin{tabular}{ll}
    \emph{Productions:} \\
    &\\
    &\\
    &\\
    &\\
    &\\
    &\\
    \end{tabular}
&
    \begin{tabular}{l}
        \emph{John} $\longrightarrow$ 12 \\
        \emph{smiles} $\longrightarrow$ 34 \\
        \emph{and} $\longrightarrow$ 13 \\
        \emph{`pronoun'} $\longrightarrow$ 1 \\
        \emph{`did too'} $\longrightarrow$ 2 \\
        \emph{EOS} $\longrightarrow$ 0
    \end{tabular}  \\
\hline 
\begin{tabular}{ll}
    \emph{Signals:} \\
    &\\
    &\\
    \end{tabular}
    & \begin{tabular}{l}
    \textbf{No Elision:} 12 34 13 12 34 0 \\
    \textbf{Pronoun:} 12 34 13 1 2 0 \\
    \textbf{Pro-drop:} 12 34 13 0 \\
    \end{tabular}\\
\end{tabular}
\end{center}
\caption{An example with corresponding signals for each handcrafted language. \emph{EOS} is the end-of-sentence token, which the Sender is required to output as the final symbol.}
\end{figure}

\begin{figure*}
\begin{center}
\includegraphics[width=\textwidth]{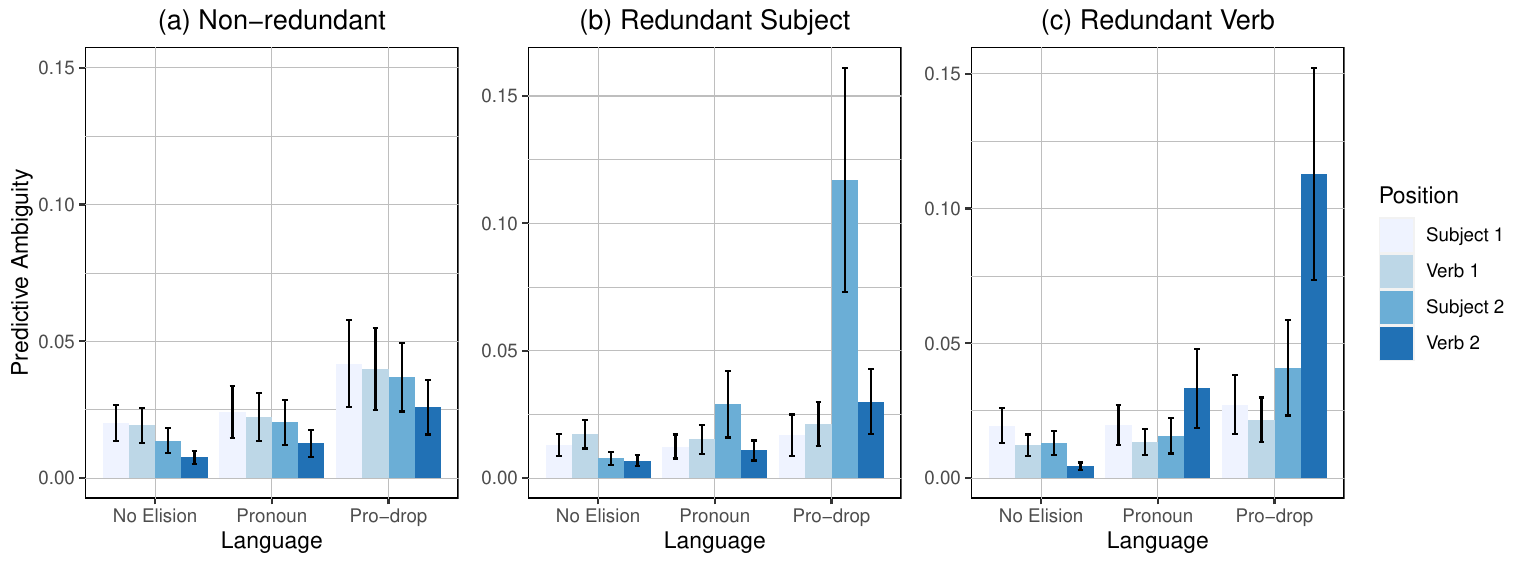}
\end{center}
\caption{Predictive ambiguity with 95\% confidence intervals for single agent experiments. Meanings are either a) non-redundant, b) partially redundant (redundant subjects), or c) partially redundant (redundant verbs) (computed at the earliest epoch when perfect test accuracy is achieved on all languages). Position 3 corresponding with \emph{and} is omitted here and in Figure 5—due to its appearance in each meaning, corresponding entropy is always low and of little interest.} 
\label{reconents1}
\end{figure*}

\subsection{Predictive Ambiguity}\label{sec:predictive_ambiguity}
Anaphoric forms like \emph{she} or \emph{did too} can be ambiguous because their intended referent is determined by context. In the next set of experiments where we look to see if these structures emerge between networks, it would be useful to quantify how ambiguous a given signal is for the Receiver. The emergence of communicatively successful, but more ambiguous, signals may be an indication that those signals contain emergent anaphors. In information-theoretic terms, more ambiguity means more uncertainty, quantifiable in terms of \emph{entropy} \citep{shannon_mathematical_1948}. The Receiver $\mathcal{R}$, when given a signal $s_i$, predicts a distribution over $words$ for each role $r \in roles$, parameterising the distribution $\mathcal{R}(words | s_i, role)$. If the Receiver is certain the first subject is Mary, then Mary will have probability 1.0, resulting in an entropy of 0.0. If the Receiver finds the meaning totally ambiguous with respect to the first subject, then we would expect that distribution over words to be uniform, resulting in higher entropy. We quantify this as \emph{predictive ambiguity} (PA), defined for each role as the average entropy over words in a role for a set of \emph{M} meanings: 

\begin{equation}\label{paeqv2}
    PA(M, role) = \frac{\sum_{i=1}^{|M|} \mathcal{H}(\mathcal{R}(words | s_i, role))}{|M|}
\end{equation}

\noindent Predictive ambiguity is shown for each of the 3 languages in Figure~\ref{reconents1}, computed separately over meanings without redundancy, redundant subjects, or redundant verbs. Redundant roles (panels (b)-(c)) show an increase in ambiguity in the Pronoun and Pro-drop languages, but no increase in the No Elision language. Conversely, for non-redundant meanings (panel (a)) ambiguity is similar for all positions and comparable across all three languages. These findings are in line with what we might expect, with anaphoric structures resulting in increased ambiguity and increased training time. They also highlight the desirability of overt anaphoric forms in a language: while serving a speaker's need for efficiency, they introduce relatively little ambiguity, allowing a listener to recover the intended meaning.

Importantly though, all three languages are learnable, meaning if a language with anaphoric structure emerged among neural agents, it could be maintained by agents exposed to it. While the languages vary in learnability—with the No Elision language the most learnable, followed by Pronoun, and finally Pro-drop—this difference is small—the Receiver only requires on average 13 additional epochs of training to successfully learn the Pro-drop language.

\section {Languages with anaphoric structure emerge between neural agents}
Having shown that neural agents reliably acquire anaphoric structure, we move now to see if such structure emerges naturally between agents, or if explicit pressures related to efficiency are required. As a proxy to effort minimisation pressures, we add an explicit term into the loss function (following \citet{chaabouni_anti-efficient_2019}) to penalise the Sender for sending longer signals:

\begin{equation}
    \centering
    \textbf{B} = \textbf{B} + \alpha \times \vert m \vert
\end{equation}

\noindent \textbf{B} represents the standard Sender loss obtained with the REINFORCE objective, while $\alpha$ is a hyperparameter, controlling how strong the efficiency pressure is, and $\vert m \vert$ is the length of the signal generated by the Sender. Across all experiments using a length cost, $\alpha = 0.15$ is used. Results are reported for two conditions: a) with length cost (\textbf{+Efficiency}), and b) without length cost (\textbf{Control}). For each condition we run 10 different initialisations, with each run having 3000 interactions between Sender and Receiver. Agents communicate about the same meaning space used in the last experiment, but here start with a random mapping from meanings to signals rather than a language of our design. We set the maximum signal length to $n = 10$ and the alphabet size to $|C| = 26$.
\begin{figure*}[h]
    \centering
    \includegraphics[width=\textwidth]{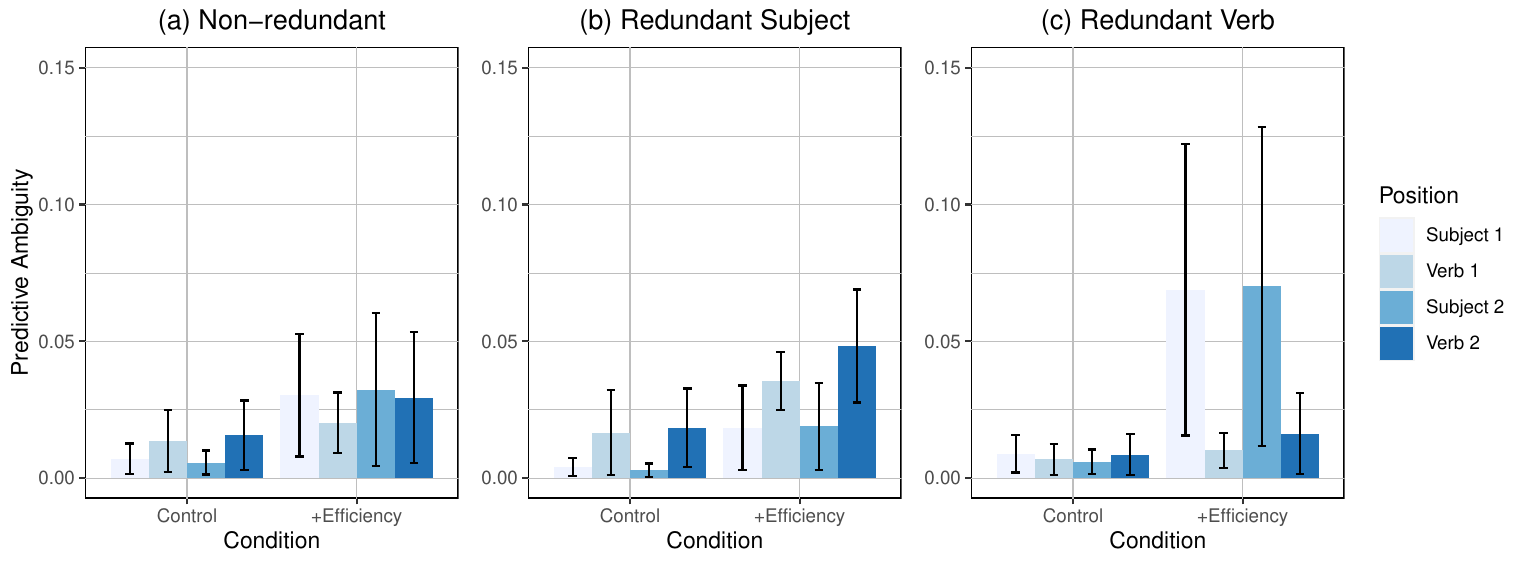}
    \caption{Predictive ambiguity in the multi-agent experiments (95\% confidence intervals).}
\label{PAEmergentLanguage}
\end{figure*}
\subsection{Identifying Anaphoric Structure}
In order to find evidence of anaphoric structure in the emergent languages, we use a set of three quantitative measurements which look for high-level characteristics of anaphoric structures in natural language. With each measurement, we compare the signals produced for different `meaning groups': those that are fully, partially, and non-redundant.

\subsubsection{Signal Uniqueness}\label{sec:signal_uniqueness}
In natural language, words used to express semantic redundancy are partially unique, since some words are used only for this purpose (e.g., \emph{she}, \emph{they}), whereas others are syntactically and semantically context-dependent (e.g., \emph{did}, \emph{too}). Emergent languages with anaphoric structure should mirror this tendency, with a subset of strings which appear only with redundant meanings—despite the fact that words in our meaning space are equally likely to appear in redundant or non-redundant contexts. We can quantify this using \emph{Jaccard similarity} \citep{jaccard1908nouvelles} to determine the overlap between \emph{n-grams} (signal substrings) used with redundant vs. non-redundant meanings. For a set of signals $S$ we can define signal uniqueness $SU$ as the difference between the Jaccard similarity for n-grams used in a sample of redundant signals $S_{red}$ and non-redundant signals $S_{\varA{non-red}}$, and a control—the Jaccard similarity for n-grams used in two random (mutually exclusive) samples of non-redundant signals $S_{\varA{non-red}}$ and $S_{\varA{non-red}'}$:

    \begin{equation}\label{jacceq}
    \centering
        SU(S) = \frac{\mid S_{\varA{non-red}} \cap S_{\varA{non-red}'} \mid}{\mid S_{\varA{non-red}} \cup S_{\varA{non-red}'} \mid} - \frac{\mid S_{red} \cap S_{\varA{non-red}} \mid}{\mid S_{red} \cup S_{\varA{non-red}} \mid}
    \end{equation}

\noindent An emergent language with anaphoric structure should have a higher $SU$ value than one without, i.e., the overlap between n-grams used in redundant and non-redundant meanings should be smaller than the overlap between two random samples of non-redundant meanings. We show in Table \ref{jacctab} that this holds for the handcrafted languages used in the preceding experiments. We compute signal uniqueness for unigrams, bigrams and trigrams in the signals respectively.

\subsubsection{Signal Length}\label{sec:signal_length}
In natural language, anaphors often shorten signals, either by fully removing redundant material as in some kinds of ellipsis (e.g., gapping) and pro-dropping, or by using short, frequent anaphoric forms. By measuring the mean length of signals for different meaning groups in a given emergent language, we can see if redundant meanings are consistently shorter, indicating the use of structures analogous to gapping or pro-drop. This may not capture instances of overt anaphoric usage given that agents may not necessarily use anaphors that are shorter than the forms they replace—we use signal uniqueness to identify anaphors independent of their length.

\subsubsection{Predictive Ambiguity}
As demonstrated in the previous section, the Receiver's predictive ambiguity is indicative of a language's anaphoric structure, with the No Elision language resulting in less predictive ambiguity than either of the languages with anaphoric structure. In the following emergent experiments, if the emergent languages do develop anaphoric structure then we would similarly expect higher predictive ambiguity—i.e., higher uncertainty—about the intended meaning for redundant roles than for non-redundant ones. An important caveat here is that higher predictive ambiguity must be coupled with high communicative success given that anaphorically structured languages are more ambiguous but still communicatively useful. A completely random language which does not enable the agents to solve the task will likely be highly ambiguous, but that in and of itself should not be taken as evidence of human-analogous anaphoric structure. Fortunately, all runs of our model achieve near perfect communicative accuracy making this issue not a concern for our results.

\subsection{Results}
All conditions achieve near-perfect communicative success on both the training data and a held-out test set. Because these results are the same across all conditions they are omitted here for brevity. We review our three measures for evidence of anaphoric structure in both the +Efficiency and Control conditions—with and without additional pressure for speaker efficiency. We find evidence that anaphoric structure emerges in all conditions. With an additional speaker-oriented pressure imposed on the system, our measures for anaphoric structure are amplified, highlighting how constraints on speaker efficiency condition the structures that emerge—even if they are not required for its emergence.

\paragraph{Signal Uniqueness} In each condition of our setup we see a unique repertoire of n-grams used to refer exclusively to redundant meanings. This is indicative of anaphoric structure of some kind having emerged, given that the model has specialised ways of conveying semantically redundant information. Additionally, this suggests that anaphoric structure emerges `naturally' without requiring any pressure for speaker brevity. Results in Table~\ref{jacctab} also highlight the amplifying effect of a length cost: for bigrams and trigrams, signal uniqueness is higher in +Efficiency. 

\begin{table}
\centering
\begin{adjustbox}{width=0.48\textwidth}
\begin{tabular}{ |*{4}{c|} }
    \hline
    & Unigram & Bigram & Trigram \\
    \hline
    No Elision & 0.0 & 0.0 & 0.0 \\
    Pronoun & 0.0 & 0.265 & 0.297 \\
    Pro-drop & 0.0 & 0.0 & 0.119 \\
    \hline
    Control & 0.0 $\pm$ 0.0 & 0.148 $\pm$ 0.0660 & 0.183 $\pm$ 0.0541 \\
    +Efficiency & 0.0 $\pm$ 0.0 & 0.210 $\pm$ 0.0397 & 0.198 $\pm$ 0.0691 \\
    \hline
\end{tabular}
\end{adjustbox}
\caption{Signal uniqueness for each condition (95\% confidence intervals). 0.0 indicates no difference in n-grams used for redundant vs. non-redundant meanings. Higher numbers indicate a larger set of n-grams are used exclusively with redundant meanings. Values for the handcrafted languages in the previous section are provided for reference. Unigrams are 0 as individual characters are used across meaning types.}
\label{jacctab}
\end{table}

\begin{table}
\centering
\begin{adjustbox}{width=0.48\textwidth}
\begin{tabular}{ |*{3}{c|} }
    \hline
    & Control & +Efficiency \\
    \hline
    All & 9.92 $\pm$ 0.041 & 9.58 $\pm$ 0.244 \\
    \hline
    Partially Redundant & 9.91 $\pm$ 0.043 & 9.51 $\pm$ 0.255 \\
    \hline
    Fully Redundant & 9.92 $\pm$ 0.041 & 9.35 $\pm$ 0.260 \\
    \hline
    Non-redundant & 9.92 $\pm$ 0.041 & 9.63 $\pm$ 0.242 \\
    \hline
\end{tabular}
\end{adjustbox}
\caption{Mean signal lengths (95\% confidence intervals).}
\label{lengthtab}
\end{table}

\paragraph{Signal Length} Overall, the mean length of all signals is very close to the maximum of 10 (see Table~\ref{lengthtab}), with lengths in the +Efficiency condition lower than for Control ($t(18)=2.47$, $p<0.05$).
Moreover, when a cost is applied, signal length decreases more for redundant than non-redundant meanings, and more so when both subject and verb are redundant, although this reduction is not significant ($t(18)=-0.683$, $p=0.503$). As such, the emergent languages for both +Efficiency and Control may develop some anaphoric structure but are unlikely to include structures analogous to elision which would more directly reduce length. In future, the prevalence of ellipsis-like structure may be increased by introducing greater efficiency costs or incentivising the model to develop other cues that can help resolve ambiguity (e.g., verbal morphology in natural languages with pro-drop may help listeners recover an antecedent when no overt pronoun is present).

\paragraph{Predictive Ambiguity} We observe a numeric trend towards increased predictive ambiguity in the redundant meanings compared with the non-redundant ones. For each role in the meaning and across different meaning groups we also see that predictive ambiguity tends to be higher for +Efficiency than for Control. These observations are not statistically significant, which could be due to high variance as a result of each run using a different network initialisation. In Figure~\ref{PAEmergentLanguage}, the redundant subject and verb meanings (panels (b)-(c)) experience predictive ambiguity `spikes' in +Efficiency—we also see evidence of this for meanings with redundant subjects in Control.\footnote{The spikes appear in the non-redundant position instead of the redundant one, e.g., we see high predictive ambiguity in Subject 2 when the verb is redundant. This may arise because signal ambiguity has an effect across the whole meaning.} These results, coupled with each run's high communicative success, further suggest that the emergent languages in all conditions contain some kind of anaphoric structure, with a speaker-oriented pressure magnifying the Receiver's uncertainty about the intended meaning for redundant roles, without degrading communicative success.

\medskip

Overall, these results provide compelling evidence that anaphoric structure emerges between neural networks. We find that, across all conditions, languages use unique n-grams that exclusively refer to redundancy and increase ambiguity about the intended meaning. These measures are further amplified when a pressure for brevity is imposed on the Sender. The minimal change in signal length with the additional pressure suggests the resulting structures resemble overt anaphors, rather than elided ones. While these results should not be interpreted as providing definitive evidence of anaphoric structure analogous to anaphora in natural language, they point towards two hypotheses about the origins of anaphoric structure: firstly, anaphoric structure does not require explicit constraints to emerge, instead emerging `for free' depending on the semantic context; secondly, pressures like efficiency condition the emergence of anaphoric structure but are not a prerequisite for it.
Encouraging the emergence of elided structures may require greater efficiency pressures, or a different semantic context.

\section{Conclusion}
In our experiments we have shown that neural agents are able to acquire languages containing anaphoric structure with ease, and that languages with overt anaphoric structure straightforwardly emerge in a communicative setting. While our evidence suggests efficiency pressures on the speaker amplify the degree of anaphoric structure—in line with expectations—strong efficiency pressures do not appear to be a precondition for its emergence. This points to the importance of the semantics-pragmatics interface, in addition to communicative needs, in offering an explanatory account of the origins of anaphoric structure.
\bibliographystyle{newapa}

\bibliography{references_v2.bib}

\end{document}